\documentclass[conference]{IEEEtran}
\IEEEoverridecommandlockouts

\usepackage{cite}
\usepackage{amsmath,amssymb}
\usepackage{graphicx}
\usepackage{booktabs}
\usepackage{array}
\usepackage{url}
\usepackage{xcolor}
\usepackage{placeins}

\graphicspath{{figures/}}
\definecolor{shared}{HTML}{2A9D8F}
\definecolor{learned}{HTML}{356FA8}
\definecolor{analytic}{HTML}{D98C32}
\definecolor{scripted}{HTML}{D9DEE3}
\definecolor{ink}{HTML}{263238}

\title{Embedded Sim-to-Real Visual Servoing: Learned Reaching and Force-Guided Placement}

\author{\IEEEauthorblockN{Dezhi Sun, Duy Khang Banh, and Kenan Niu\textsuperscript{*}}
\thanks{The authors are with Robotics and Mechatronics, Faculty of Electrical Engineering, Mathematics and Computer Science, University of Twente, Enschede, The Netherlands.\protect\newline
\textsuperscript{*}Corresponding author: Kenan Niu.\protect\newline \textcopyright~2026 IEEE. Personal use of this material is permitted. Permission from IEEE must be obtained for all other uses, in any current or future media, including reprinting/republishing this material for advertising or promotional purposes, creating new collective works, for resale or redistribution to servers or lists, or reuse of any copyrighted component of this work in other works.}}

\begin{document}
\maketitle

\begin{abstract}
Embedded perception-to-control systems must combine visual targeting
with reliable contact handling under tight computational constraints.
We present a shared architecture on an NVIDIA Jetson AGX Orin
controlling a Franka Research~3 through Cartesian impedance control.
RealSense D435i observations, YOLO11-seg segmentation, depth
back-projection, and hand--eye calibration provide metric robot-frame
targets. We compare two strategies: compact proximal policy
optimization (PPO) policies transferred from simulation for object-
and hole-region reaching, and force-guided Cartesian placement
using approximately 2~N normal contact and spiral search.
The PPO policies achieved 87.2\% and 83.1\% success over 1000
simulation episodes each. Across 50 real-robot trials per strategy,
the learned hybrid and force-guided analytical pipelines achieved
70.0\% and 74.0\% completion, respectively.
Mean perception latency with TensorRT was 34.55~ms.
These results support an embedded architecture combining learned
free-space reaching with explicit contact correction.
\end{abstract}

\begin{IEEEkeywords}
Embedded robotics, visual servoing, sim-to-real transfer, reinforcement learning, robotic manipulation, force control, RGB-D perception.
\end{IEEEkeywords}

\section{Introduction}
\label{sec:introduction}

Embedded visual manipulation integrates perception, motion generation, and physical interaction under computational constraints. RGB-D sensing supports free-space targeting, but constrained placement remains sensitive to localization and orientation errors. Robotic imaging combines visual perception with contact-force regulation~\cite{sun2025arus,sun2025hcce} and coordinated motion control~\cite{ward2025bilateral}. Contact and geometric jamming further require compliant local correction~\cite{Whitney1982,Xu2019}.

Visual feedback supports robotic control~\cite{Hutchinson1996,Chaumette2006,zhang2026bci}, while learning-based manipulation acquires target-conditioned behavior from interaction~\cite{Levine2016,Kalashnikov2018}. RLBench provides simulation-based training~\cite{James2020}, and domain or dynamics randomization facilitates sim-to-real transfer~\cite{Tobin2017,Peng2018}. Impedance control, spiral search, and residual learning address contact-rich placement errors~\cite{Hogan1985,Park2020,Johannink2019,Ankile2024}, with related compliant-control applications in robotic catheterization~\cite{wu2022compliant} and ultrasound scanning~\cite{davoodi2023control}.

Large-scale visuomotor policies offer broad manipulation capabilities but typically require substantial models and robot data~\cite{Brohan2023RT1,Kim2024OpenVLA,Chi2023Diffusion}. We instead evaluate compact state-based PPO reaching and force-guided analytical placement on a shared embedded platform. YOLO11-seg processes eye-in-hand RGB-D observations on a Jetson AGX Orin to reconstruct targets in the Franka Research~3 (FR3) base frame. Both strategies share sensing, calibration, hardware, and Cartesian impedance execution; the analytical branch additionally uses force and end-effector height during placement.

The main contributions are: (1) \textbf{Embedded architecture---}a Jetson-based pipeline integrating RGB-D segmentation, metric target reconstruction, and compliant FR3 execution; (2) \textbf{Matched comparison---}compact sim-to-real PPO reaching and force-guided analytical placement under common sensing and execution conditions; and (3) \textbf{Experimental validation---}1000 simulation episodes per policy and 50 real-robot trials per strategy. Physical manipulation evaluates an approximately aligned square object, while perception covers all four classes.
\begin{figure*}[t]
  \centering
  \includegraphics[width=0.98\textwidth]{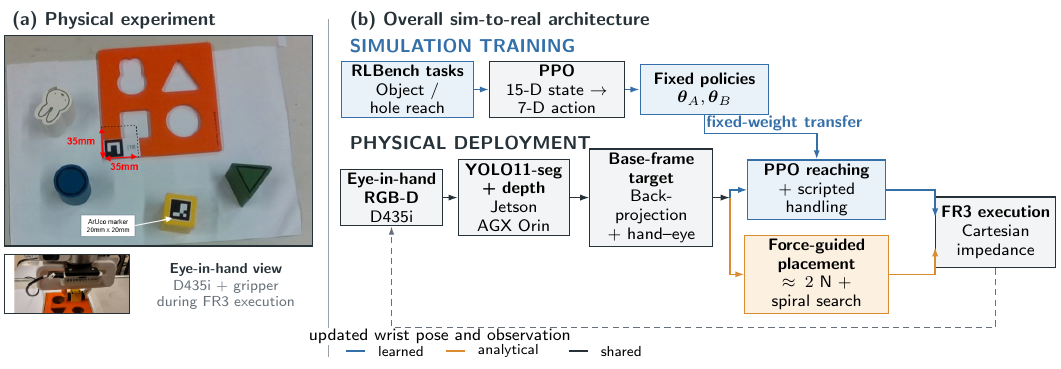}
  \caption{Physical experiment and overall sim-to-real architecture. (a) Authentic views of the sorting workspace and wrist-mounted D435i/gripper during FR3 execution. (b) Policy~A and Policy~B are trained in RLBench and transferred with fixed weights. During deployment, Jetson-based RGB-D perception reconstructs a common base-frame target for either PPO reaching or force-guided placement, both executed through Cartesian impedance control.}
  \label{fig:architecture}
\end{figure*}

\section{System and Methods}
\label{sec:methods}

The system separates target reconstruction from motion generation. A shared embedded front end produces a class-labelled metric target in the FR3 base frame; either motion branch converts it into commands tracked by the same compliant execution layer. This preserves the distinct roles of learned reaching and analytical contact correction.

\subsection{Shared Embedded Perception and Control}

Figure~\ref{fig:architecture} connects RLBench policy learning to physical deployment. The transferred policies retain fixed weights, while the real system reconstructs the simulator-compatible metric target from eye-in-hand RGB-D observations. Perception and high-level motion generation execute on the Jetson AGX Orin. ROS1, \texttt{franka\_ros}, and \texttt{libfranka} connect target estimates to compliant FR3 control~\cite{Quigley2009,Chitta2017}; TensorRT provides embedded YOLO11-seg inference~\cite{Mittal2019,Ultralytics2024}.

The YOLO11-seg model processes color images at input size 736. It was trained from approximately 100 labeled RGB-D images using an 80/20 training--validation split and represents four classes: Blue circle, Green triangle, Nijntje/Miffy, and Yellow square. Aligned depth is sampled at each mask centroid, with invalid or near-zero values rejected. For centroid $(u,v)$, depth $z$, and camera intrinsics $(f_x,f_y,c_x,c_y)$, the homogeneous camera-frame point is
\begin{equation}
\mathbf{p}_{C}=
\begin{bmatrix}
(u-c_x)z/f_x & (v-c_y)z/f_y & z & 1
\end{bmatrix}^{\!T}.
\label{eq:backprojection}
\end{equation}
Here, $u$ and $v$ are image coordinates, $z$ is metric depth, and $f_x,f_y,c_x,c_y$ are the calibrated focal lengths and principal point.

Tsai--Lenz eye-in-hand calibration~\cite{Tsai1989} and ArUco references~\cite{Garrido2014} establish the camera-to-robot geometry. Combining the rigid hand--eye transform with current FR3 kinematics yields $\mathbf{T}^{B}_{C}\in\mathrm{SE}(3)$, giving
\begin{equation}
\mathbf{p}_{B}=\mathbf{T}^{B}_{C}\mathbf{p}_{C}, \qquad
\mathbf{p}_{B}(1{:}3)=[x_B,y_B,z_B]^T,
\label{eq:base_transform}
\end{equation}
where $\mathbf{p}_{B}$ is the target in the FR3 base frame. The ROS \texttt{tf} tree combines the fixed calibration with the instantaneous eye-in-hand pose, so updated wrist-camera observations produce updated metric targets.

This metric target is the interface between perception and motion generation. The learned branch inserts it into a state observation; the analytical branch constructs Cartesian approach and placement poses. Both are tracked through Cartesian impedance control, keeping compliance and low-level execution fixed.

\begin{table}[t]
\caption{Embedded architecture and compact policy configuration.}
\label{tab:configuration}
\centering
\renewcommand{\arraystretch}{1.03}
\setlength{\tabcolsep}{2.6pt}
\begin{tabular}{@{}p{0.28\columnwidth}p{0.65\columnwidth}@{}}
\toprule
\textbf{Item} & \textbf{Configuration} \\
\midrule
\multicolumn{2}{@{}l}{\textbf{Hardware}} \\
Robot / control & Franka Research~3 / Cartesian impedance \\
Camera & Intel RealSense D435i \\
Embedded compute & NVIDIA Jetson AGX Orin \\
\addlinespace[1pt]\multicolumn{2}{@{}l}{\textbf{Perception}} \\
Detector / format & YOLO11-seg / TensorRT \\
Input resolution & 736 \\
\addlinespace[1pt]\multicolumn{2}{@{}l}{\textbf{Policy}} \\
Policy & PPO MLP with \texttt{Tanh} activations \\
Observation / action & 15 state values / 7 joint increments \\
Actor / critic & 15--64--64--7 / 15--64--64--1 \\
Policy size & \textbf{10,895 parameters; $\sim$43~kB FP32} \\
\addlinespace[1pt]\multicolumn{2}{@{}l}{\textbf{Training}} \\
Simulation & RLBench / PyRep / CoppeliaSim \\
Environment steps & \textbf{$10^7$ per policy} \\
Parallel environments & \textbf{16} \\
\bottomrule
\end{tabular}
\end{table}

\begin{figure*}[!t]
  \centering
  \includegraphics[width=0.98\textwidth]{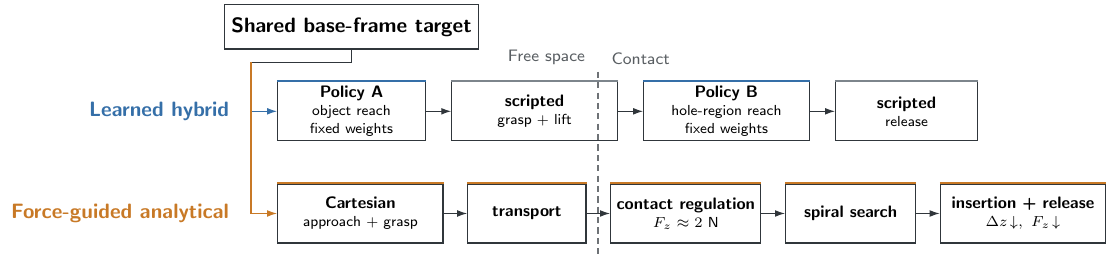}
  \caption{Motion-stage decomposition. The learned hybrid uses fixed-weight PPO policies for object- and hole-region reaching, with scripted handling between the two stages. The analytical pipeline adds normal-force regulation and spiral search after contact. Blue and orange lanes identify learned and analytical modules, respectively.}
  \label{fig:pipelines}
\end{figure*}

\subsection{Sim-to-Real Learned Reaching Primitives}

Policy~A learns object-region reaching and Policy~B learns hole-region reaching in RLBench~\cite{James2020}, using PyRep~\cite{James2019} and CoppeliaSim~\cite{Rohmer2013}. Each PPO policy~\cite{Schulman2017} was trained for 10 million environment steps with 16 parallel environments. Training used 2048-step rollouts, mini-batches of 64, 10 optimization epochs per update, a $3{\times}10^{-4}$ learning rate, and a 0.1 clip range. The 15-dimensional observation contains seven joint positions, three-dimensional end-effector position, an object-class indicator, three-dimensional target position, and one orientation term. Seven incremental joint commands form the action. Actor and critic networks use two 64-unit hidden layers with \texttt{Tanh} activations, yielding the architectures and storage footprint in Table~\ref{tab:configuration}. Rewards combine distance progress, target-proximal orientation/alignment shaping, terminal success, and collision penalty.

Policy~A targets the object region and Policy~B the hole region. Policy~B was trained as unloaded reaching, without a grasped object or contact-rich insertion; its role is therefore coarse hole-region approach. The actors output continuous joint increments, while the critics estimate state value during training; image inference remains outside the compact policies.

For sim-to-real transfer, the robot reconstructs a 15-dimensional observation with the same semantics, units, ordering, and coordinate frames used in simulation. Joint states come from the FR3 interface, end-effector pose from kinematics and \texttt{tf}, and metric targets from Eqs.~\eqref{eq:backprojection}--\eqref{eq:base_transform}. The fixed policy weights thus consume calibrated state vectors rather than raw images, without real-world fine-tuning.

Figure~\ref{fig:pipelines} shows how deterministic task primitives connect the two fixed-weight policies. The simulation success thresholds are 0.010~m for Policy~A and 0.005~m for Policy~B. Policy~A supplies object-region reaching, while Policy~B supplies the coarse hole-region approach.

\subsection{Force-Guided Analytical Placement}

The force-guided analytical strategy executes deterministic Cartesian approach, grasp, transport, contact-guided placement, release, and retraction through the same compliant interface. The vision-derived targets define approach and transport above the sorter. The controller then descends to approximately 2~N normal contact and applies the planar spiral
\begin{equation}
\Delta x(t) = a t \cos(2\pi f t), \qquad
\Delta y(t) = a t \sin(2\pi f t).
\label{eq:spiral}
\end{equation}
Here, $a$ sets radial growth and $f$ the rotation frequency. The offsets explore the neighborhood around the vision-derived hole center while Cartesian impedance control maintains contact.

Entry permits the square and end effector to descend while force relaxes, distinguishing insertion from surface sliding. The controller detects this coupled height--force change, reduces stiffness, releases, and retracts. Vision supplies the global target; interaction supplies local evidence of entry.

\subsection{Real-Robot Implementation}

The eye-in-hand target is updated online rather than fixed at the start of a trial. At each control update, the perception node combines the current D435i color and aligned-depth images with the rigid hand--eye calibration and instantaneous FR3 kinematics. ROS joint states, the end-effector transform, the object-class indicator, and the reconstructed target then populate the same 15-dimensional observation used in simulation. This state-based interface keeps image processing outside the policy and permits the two fixed-weight PPO actors to run without real-world policy fine-tuning.

The learned branch produces seven incremental joint commands at approximately 30~Hz. Each increment is clamped before execution, converted through forward kinematics into an end-effector equilibrium target, and tracked by Cartesian impedance control. Policy~A terminates within 0.014~m of the physical object target; scripted grasp and lift then prepare the second stage. Policy~B terminates within 0.006~m of the physical hole-region target, after which release and retraction remain deterministic. These deployment thresholds preserve the relative precision of the 0.010 and 0.005~m simulation criteria while accommodating sensing and calibration residuals.
The resulting increases are 4~mm for Policy~A and 1~mm for Policy~B, so the physical implementation relaxes both stopping tests without reversing their relative strictness.

The analytical branch uses the same visual target for Cartesian approach, grasp, and transport. Placement begins after approximately 2~N normal contact is established. The controller then superimposes the planar spiral offsets in Eq.~\eqref{eq:spiral} while monitoring end-effector height and normal force. A coupled downward displacement and force relaxation indicates entry; stiffness is reduced before release and safe retraction. Thus both branches share perception and compliant execution, but only the analytical branch closes a local loop on contact during final placement.

\begin{figure*}[!t]
  \centering
  \includegraphics[width=0.98\textwidth]{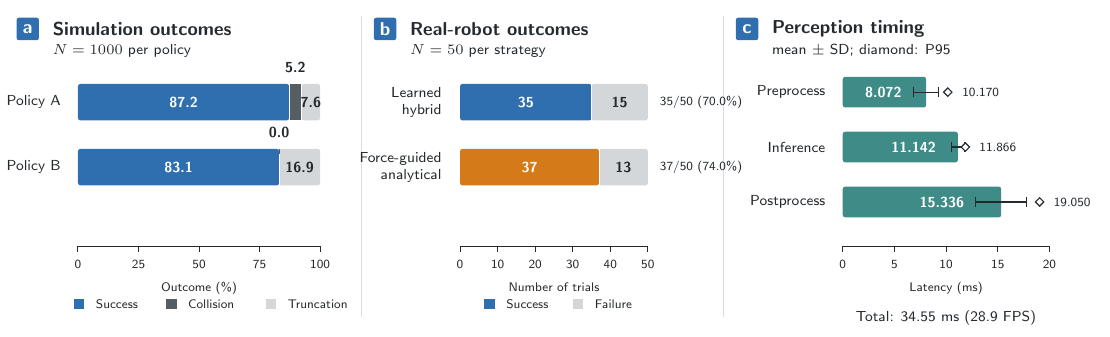}
  \caption{Quantitative evaluation. (a) Simulation terminal outcomes over 1000 episodes per policy. (b) End-to-end task outcomes over 50 physical trials per strategy. (c) Mean and P95 TensorRT processing latency over 1000 frames; error bars denote standard deviation.}
  \label{fig:quantitative}
\end{figure*}

\section{Experimental Protocol}
\label{sec:protocol}

\subsection{Embedded Perception}

Approximately 100 D435i RGB-D images spanning four segmentation classes were divided 80/20 for training and validation. The validation set comprised 20 images and contained 20 annotated instances per class. Precision, recall, mAP50, and mAP50-95 quantify class-wise masks; TensorRT latency was measured over 1000 Jetson frames.

\subsection{Simulation}

Each policy was evaluated over $N=1000$ RLBench episodes. Success, collision, and truncation are mutually exclusive: success meets the policy-specific target criterion, collision records contact termination, and truncation denotes horizon expiry.

\subsection{Matched Real-Robot Conditions}

Fifty learned hybrid and 50 force-guided analytical trials used the same FR3, D435i, Jetson stack, calibration, square object, sorter, gripper, workspace, and impedance interface. The sorter and object were approximately aligned; the comparison therefore concerns motion generation under matched sensing and execution rather than orientation generalization.

\subsection{Real-Robot Success Criteria}

A learned hybrid trial succeeds only when Policy~A reaches the object target, the scripted grasp and lift complete, Policy~B reaches the target-hole region, the object is released through the correct opening, and the sequence terminates without an abort. This definition evaluates the complete learned-plus-scripted pipeline; reaching a target alone is not counted as task success.

An analytical trial succeeds only when Cartesian approach and grasp complete, the object is transported to the target region, controlled contact is established, spiral-guided entry is detected, the object is released through the opening, and the robot retracts safely. The two definitions use branch-specific internal stages but the same end-to-end requirement: successful object acquisition, transport, placement, release, and safe completion.

\section{Results}
\label{sec:results}

\subsection{Embedded Perception}

Table~\ref{tab:perception} reports the four segmentation classes. Macro precision was 0.987, recall 0.844, mAP50 0.953, and mAP50-95 0.791. Green triangle produced the lowest recall (0.7500) and mAP50-95 (0.6627).

\begin{table}[t]
\caption{Four-class segmentation performance in the sorting workspace.}
\label{tab:perception}
\centering
\renewcommand{\arraystretch}{1.06}
\setlength{\tabcolsep}{2.8pt}
\begin{tabular}{@{}lcccc@{}}
\toprule
\textbf{Class} & \textbf{P} & \textbf{R} & \textbf{mAP50} & \textbf{mAP50-95} \\
\midrule
Blue circle & 0.9613 & 0.8500 & 0.9360 & 0.8056 \\
Green triangle & 0.9858 & 0.7500 & 0.9160 & 0.6627 \\
Nijntje/Miffy & 1.0000 & 0.8478 & 0.9855 & 0.8184 \\
Yellow square & 1.0000 & 0.9270 & 0.9736 & 0.8786 \\
\midrule
\textbf{Macro} & \textbf{0.987} & \textbf{0.844} & \textbf{0.953} & \textbf{0.791} \\
\bottomrule
\end{tabular}
\end{table}

Mean preprocessing, inference, and postprocessing latencies were $8.072\pm1.221$, $11.142\pm0.646$, and $15.336\pm2.469$~ms; corresponding 95th percentiles were 10.170, 11.866, and 19.050~ms. Total mean latency was 34.550~ms, or approximately 28.9~FPS. This is perception throughput only; robot timing additionally includes state acquisition, target transformation, and compliant execution.

\subsection{Simulation Evaluation}

Figure~\ref{fig:quantitative}(a) summarizes the $N=1000$ evaluations. Policy~A achieved 872 successes (87.2\%), 52 collisions (5.2\%), and 76 truncations (7.6\%); Policy~B achieved 831 successes (83.1\%), 0 observed collisions (0.0\%), and 169 truncations (16.9\%). Episode lengths were $80.44\pm40.13$ and $90.95\pm49.99$ steps, with throughputs of $81.04\pm3.69$ and $85.88\pm2.37$~FPS, respectively.

Policy~A converted more episodes to success but occasionally contacted the environment. Under its tighter 0.005~m hole-region criterion, Policy~B recorded no collision and more horizon truncations; Policy~A used a 0.010~m criterion. These stage-level simulation outcomes are distinct from end-to-end physical sorting.

\subsection{Real-Robot Evaluation}

Across 100 physical executions, the learned hybrid pipeline completed 35/50 trials (70.0\%) and the force-guided analytical pipeline 37/50 (74.0\%). Wilson 95\% confidence intervals were $[56.2,80.9]\%$ and $[60.4,84.1]\%$, respectively. The overlapping confidence intervals support comparable aggregate performance in the evaluated setup. Table~\ref{tab:strategy_comparison} isolates the branch-specific components under the matched sensing and execution conditions.

\begin{table*}[t]
\caption{Controller composition under matched perception and compliant execution.}
\label{tab:strategy_comparison}
\centering
\renewcommand{\arraystretch}{1.08}
\setlength{\tabcolsep}{5.0pt}
\begin{tabular}{@{}p{0.19\textwidth}p{0.37\textwidth}p{0.37\textwidth}@{}}
\toprule
\textbf{Property} & \textbf{Learned hybrid} & \textbf{Force-guided analytical} \\
\midrule
Shared input & Class-labelled target in the FR3 base frame & Class-labelled target in the FR3 base frame \\
Motion representation & Incremental joint commands mapped to Cartesian equilibrium targets & Deterministic Cartesian equilibrium targets \\
Learned components & Policy~A object reach; Policy~B hole-region reach & None \\
Scripted components & Grasp, lift, release, retraction & Approach, grasp, transport, release, retraction \\
Contact feedback & Not included in policy state & Normal force and end-effector height \\
Local correction & No explicit insertion-state correction & Force regulation and bounded spiral search \\
Real-robot trials & 50 & 50 \\
End-to-end success & 35/50 (70.0\%) & 37/50 (74.0\%) \\
\bottomrule
\end{tabular}
\end{table*}

\subsection{System Failure Mechanisms}

Three coupled vulnerabilities were observed. First, the wrist-mounted gripper can occlude the square near grasp; because the camera and robot continue moving, target age and confidence should pause motion when the last estimate becomes stale. Second, false positives can switch the selected instance and create discontinuous targets. Class persistence, workspace bounds, and a maximum update rate would gate both branches against these jumps.

Third, aligned-depth noise, residual hand--eye calibration error, and unstable orientation estimates perturb placement. Contact search can correct lateral target bias but not arbitrary orientation error, while the learned branch lacks equivalent insertion-state evidence. These are representative mechanisms, not measured failure frequencies.

\section{Discussion}
\label{sec:discussion}

\subsection{Representation and Transfer Interface}

The 15-dimensional interface externalizes appearance from the learned controller: segmentation, aligned depth, calibration, and kinematics convert pixels into the same FR3 base-frame target used in simulation. The actor sees joint and end-effector state, object identity, and target coordinates rather than images. This reduces visual-domain shift without removing calibration, dynamics, friction, or execution discrepancies, while retaining a 10,895-parameter actor--critic pair (approximately 43~kB FP32).

Equation~\eqref{eq:backprojection} exposes how sensing error reaches either controller. A centroid perturbation creates lateral camera-frame error proportional to $z/f_x$ and $z/f_y$, whereas a depth perturbation affects the lateral coordinates through the normalized image location and affects $z$ directly. The rigid hand--eye transform rotates these errors into the base frame but cannot attenuate them. Target validation should therefore consider metric displacement after back-projection, not detector confidence alone.

Uncertainty is concentrated at the reconstructed target, so an instance-preserving gate should reject depth outliers and motion-inconsistent jumps before either controller receives them. Physical radii of 0.014 and 0.006~m add tolerance to the 0.010 and 0.005~m simulation criteria while keeping Policy~B tighter. Its 16.9\% truncation rate, versus 7.6\% for Policy~A, is consistent with that stricter terminal radius.

\subsection{Execution and Embedded Timing}

After clamping, forward kinematics converts each seven-joint increment into an end-effector equilibrium target tracked by Cartesian impedance. This preserves incremental policy motion while keeping compliant low-level execution common to both branches.

Postprocessing (15.336~ms) exceeds TensorRT inference (11.142~ms) because mask decoding, centroid extraction, and depth association are included. It accounts for 44.4\% of the 34.550~ms perception path, compared with 32.2\% for inference and 23.4\% for preprocessing. Consequently, accelerating the network alone would not remove the largest measured latency component; even zero inference time leaves 23.408~ms of measured processing, corresponding to an upper bound of approximately 42.7 updates per second unless preprocessing or postprocessing also changes. The measured total supports approximately 28.9 perception updates per second but excludes state acquisition, frame transformation, command construction, and robot execution. Postprocessing and synchronized ROS/controller timestamps should therefore be profiled before making end-to-end latency claims.

\subsection{Information Gained at Contact}

A base-frame visual target describes displacement but not sliding, jamming, or entry. The analytical branch adds this missing evidence after contact: approximately 2~N normal force maintains engagement, a bounded spiral corrects lateral bias, and decreasing height with relaxing force triggers release. Policy~B instead receives neither force nor insertion state and therefore supplies only coarse hole-region reaching before scripted release.

A contact-aware learned stage would require force/height history, contact-rich training, switch hysteresis, and minimum dwell time. The matched outcomes highlight the different feedback mechanisms used in free-space and contact-rich motion. These observations motivate a regime-aware architecture that transitions from learned free-space motion to force/height correction at contact~\cite{Chaumette2007,Johannink2019,Ankile2024}.

\subsection{Interpretation of the Matched Comparison}

The matched protocol controls the hardware, reconstructed target, workspace, object, and compliant execution interface, but it does not make the internal task decompositions identical. Learned-hybrid success requires both fixed-weight reaching stages and the intervening scripted handling to complete; analytical success requires the Cartesian sequence and contact-guided placement to complete. The strategies are compared at the common end-to-end system boundary rather than as interchangeable low-level controllers.

With 50 trials per strategy, one changed outcome shifts a reported rate by two percentage points, and the observed four-point separation corresponds to two executions. The overlapping Wilson intervals are consistent with comparable task-level performance under the matched conditions. Policy~B's 83.1\% simulation result corresponds to unloaded hole-region reaching under its stage-specific criterion, whereas physical placement is evaluated end to end.

\subsection{Scope and Limitations}

The physical study covers one approximately aligned square-object condition and aggregate pipeline completion. Orientation-aware manipulation of the other detected classes, controlled perturbations, temporal target filtering, and stage-resolved failure labels remain future work. These extensions are needed to separate perception, calibration, grasp, and insertion contributions and to test whether contact-aware policy inputs improve placement.

\section{Conclusion}
\label{sec:conclusion}

This paper presented an embedded RGB-D manipulation architecture combining compact sim-to-real PPO reaching with force-guided placement. Across 50 real-robot trials per strategy, the learned hybrid and analytical pipelines achieved 70.0\% and 74.0\% success, respectively. These results motivate combining learned free-space reaching with explicit contact correction on a shared embedded platform.

\FloatBarrier
\bibliographystyle{IEEEtran}
\bibliography{references}

\end{document}